\documentclass{article}

\usepackage{graphicx,amscd,amsmath,amssymb,verbatim}
\usepackage[dvips]{hyperref}
\usepackage[TS1,OT1,T1]{fontenc}

\begin{document}

\title{Using Stochastic Encoders to Discover Structure in Data\footnote{Full
version of a short paper that was published in the Digest of the
5th IMA International Conference on Mathematics in Signal Processing,
18-20 December 2000, Warwick University, UK.}}
\author{Stephen Luttrell}
\maketitle

\noindent {\bfseries Abstract:} In this paper a stochastic generalisation
of the standard Linde-Buzo-Gray (LBG) approach to vector quantiser
(VQ) design is presented, in which the encoder is implemented as
the sampling of a vector of code indices from a probability distribution
derived from the input vector, and the decoder is implemented as
a superposition of reconstruction vectors. This stochastic VQ (SVQ)
is optimised using a minimum mean Euclidean reconstruction distortion
criterion, as in the LBG case. Numerical simulations are used to
demonstrate how this leads to self-organisation of the SVQ, where
different stochastically sampled code indices become associated
with different input subspaces.
\section{Introduction}

In vector quantisation a code book is used to encode each input
vector as a corresponding code index, which is then decoded (again,
using the codebook) to produce an approximate reconstruction of
the original input vector \cite{{Gray1984, GershoGray1992}}. The
purpose of this paper is to generalise the standard approach to
vector quantiser (VQ) design \cite{LindeBuzoGray1980}, so that each
input vector is encoded as a vector of code indices that are stochastically
sampled from a probability distribution that depends on the input
vector, rather than as a single code index that is the deterministic
outcome of finding which entry in a code book is closest to the
input vector. This will be called a stochastic VQ (SVQ), and it
includes the standard VQ as a special case. Note that this approach
is different from the various stochastic approches that are used
to train VQs (see e.g. \cite{{ZegerGersho1989, YairZegerGersho1992,
TorresCasasArias1997}}), because here the codebook itself is stochastic,
so the use of probability distributions is essential both during
and after training.

One advantage of using the stochastic approach, which will be demonstrated
in this paper, is that it automates the process of splitting high-dimensional
input vectors into low-dimensional blocks before encoding them,
because minimising the mean Euclidean reconstruction error can encourage
different stochastically sampled code indices to become associated
with different input subspaces \cite{Luttrell1999a}. Another advantage
is that it is very easy to connect SVQs together, by using the vector
of code index probabilities computed by one SVQ as the input vector
to another SVQ \cite{Luttrell1999b}.

In Section \ref{XRef-Section-821163549} various pieces of previously
published theory are unified to give a coherent account of SVQs.
In Section \ref{XRef-Section-8211742} the results of some new numerical
simulations are presented, which demonstrate how the code indices
in a SVQ can become associated in various ways with input subspaces.
In the appendices various derivations relating to the detailed training
of an SVQ are presented.
\section{Theory}\label{XRef-Section-821163549}

In this section various pieces of previously published theory are
unified to establish a coherent framework for modelling SVQs. In
Section \ref{XRef-Subsection-821163723} the basic theory of folded
Markov chains (FMC) is given \cite{Luttrell1994}, and in Section
\ref{XRef-Subsection-821163735} it is extended to the case of high-dimensional
input data \cite{Luttrell1997}. Finally, in\ \ Section \ref{XRef-Subsection-821163750}
the theory is further generalised to chains of linked FMCs \cite{Luttrell1999b}.
\subsection{Folded Markov Chains}\label{XRef-Subsection-821163723}

\begin{figure}[h]
\begin{center}
\includegraphics{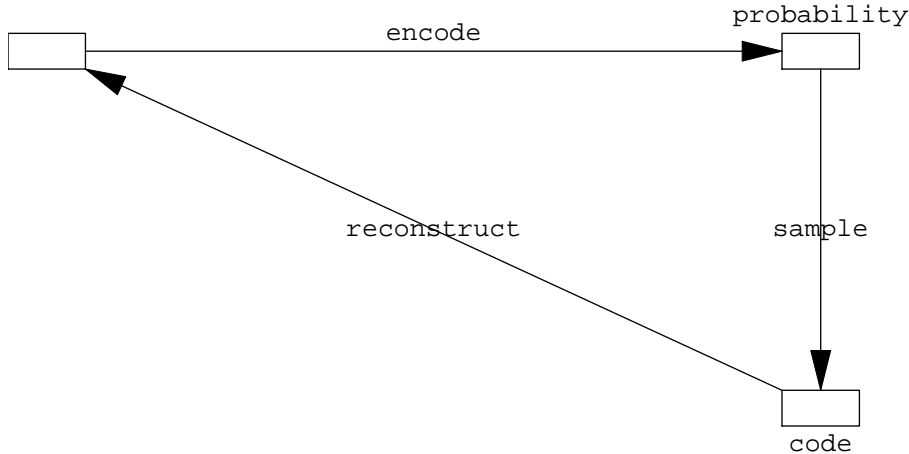}

\end{center}
\caption{A folded Markov chain (FMC) in which an input vector $x$
is encoded as a code index vector $y$ that is drawn from a conditional
probability $\Pr ( y|x) $, which is then decoded as a reconstruction
vector $x^{\prime }$ drawn from the Bayes' inverse conditional probability
$\Pr ( x^{\prime }|y) $.}\label{XRef-FigureCaption-82116438}
\end{figure}

The basic building block of the encoder/decoder model used in this
paper is the folded Markov chain (FMC) \cite{Luttrell1994}. Thus
an input vector $x$ is encoded as a code index vector $y$, which
is then subsequently decoded as a reconstruction $x^{\prime }$ of
the input vector. Both the encoding and decoding operations are
allowed to be probabilistic, in the sense that $y$ is a sample drawn
from $\Pr ( y|x) $, and $x^{\prime }$ is a sample drawn from $\Pr
( x^{\prime }|y) $, where $\Pr ( y|x) $ and $\Pr ( x^{\prime }|y)
$ are Bayes' inverses of each other, as given by $\Pr ( x^{\prime
}|y) =\frac{\Pr ( y|x) \Pr ( x) }{\int dz \Pr ( y|z) \Pr ( z) }$,
and $\Pr ( x) $ is the prior probability from which $x$ was sampled.
Because the chain of dependences in passing from $x$ to $y$ and
then to $x^{\prime }$ is first order Markov (i.e. it is described
by the directed graph ($x\longrightarrow y\longrightarrow x^{\prime
}$), and because the two ends of this Markov chain (i.e. $x$ and
$x^{\prime }$) live in the same vector space, it is called a {\itshape
folded} Markov chain (FMC). The operations that occur in an FMC
are summarised in Figure \ref{XRef-FigureCaption-82116438}.

In order to ensure that the FMC encodes the input vector optimally,
a measure of the reconstruction error must be minimised. There are
many possible ways to define this measure, but one that is consistent
with many previous results, and which also leads to many new results,
is the mean Euclidean reconstruction error measure $D$, which is
defined as
\begin{equation}
D\equiv \int dx \Pr ( x) \sum \limits_{y_{1}=1}^{M}\sum \limits_{y_{2}=1}^{M}\cdots
\sum \limits_{y_{n}=1}^{M}\Pr ( y|x) \int dx^{\prime }\Pr ( x^{\prime
}|y) \left| \left| x-x^{\prime }\right| \right| ^{2}
\end{equation}

\noindent where $y=(y_{1},y_{2},\cdots ,y_{n}), 1\leq y_{i}\leq
M$ is assumed, $\Pr ( x) \Pr ( y|x) \Pr ( x^{\prime }|y) $ is the
joint probability that the FMC has state $(x,y,x^{\prime })$, $||x-x^{\prime
}||^{2}$ is the Euclidean reconstruction error, and $\int dx \sum
\limits_{y_{1}=1}^{M}\sum \limits_{y_{2}=1}^{M}\cdots \sum \limits_{y_{n}=1}^{M}\int
dx^{\prime }( \cdots ) $ sums over all possible states of the FMC
(weighted by the joint probability).

The Bayes' inverse probability $\Pr ( x^{\prime }|y) $ may be integrated
out of this expression for $D$ to yield
\begin{equation}
D=2\int dx \Pr ( x) \sum \limits_{y_{1}=1}^{M}\sum \limits_{y_{2}=1}^{M}\cdots
\sum \limits_{y_{n}=1}^{M}\Pr ( y|x) \left| \left| x-x^{\prime }(
y) \right| \right| ^{2}
\end{equation}

\noindent where the reconstruction vector $x^{\prime }( y) $ is
defined as $x^{\prime }( y) \equiv \int dx \Pr ( x|y) x$. Because
of the quadratic form of the objective function, it turns out that
$x^{\prime }( y) $ may be treated as a free parameter whose optimum
value (i.e. the solution of $\frac{\partial D}{\partial x^{\prime
}( y) }=0$) is $\int dx \Pr ( x) x$, as required.

It was shown in \cite{Luttrell1994} that the standard VQ \cite{LindeBuzoGray1980}
and topograpic mappings \cite{Kohonen1984} automatically emerge
as special cases when $D$ is minimised. In this approach, topographic
mappings emerge as the optimal coding scheme when the code is to
be transmitted along a noisy communication channel before being
decoded \cite{{Farvardin1990, KumazawaKasaharaNamekawa1984}}.
\subsection{High Dimensional Input Spaces}\label{XRef-Subsection-821163735}

A problem with the standard VQ is that its code book grows exponentially
in size as the dimensionality of the input vector is increased,
assuming that the contribution to the reconstruction error from
each input dimension is held constant. This means that such VQs
are useless for encoding extremely high dimensional input vectors,
such as images. The usual solution to this problem is to manually
partition the input space into a number of lower dimensional subspaces,
and then to encode each of these subspaces separately. However,
it would be very useful if this partitioning could be done automatically,
in such a way that typically the correlations {\itshape within}
each subspace were much stronger than the correlations {\itshape
between} subspaces, so that the subspaces were approximately statistically
independent of each other. The purpose of this paper is to present
a solution to this problem. 

The key step in solving this problem is to constrain the minimisation
of $D$ in such a way as to encourage the formation of code schemes
in which each component of the code vector $y$ codes a different
subspace of the input vector $x$. There are two related constraints
that may be imposed on $\Pr ( y|x) $ and $x^{\prime }( y) $ which
may be summarised as
\begin{equation}
\begin{split}
\Pr ( y|x) &=\Pr ( y_{1}|x) \Pr ( y_{2}|x) \cdots  \Pr ( y_{n}|x)
\\
x^{\prime }( y) &=\frac{1}{n}\sum \limits_{i=1}^{n}x^{\prime }(
y_{i}) 
\end{split}%
\label{XRef-Equation-821165127}
\end{equation}

\noindent Thus each component $y_{i}$ (for $i=1,2,\cdots ,n$ and
$1\leq y_{i}\leq M$) is an {\itshape independent} sample drawn from
the codebook using $\Pr ( y_{i}|x) $ (which is assumed to be the
same function for all $i$), and the reconstruction vector $x^{\prime
}( y) $ (vector argument) is assumed to be a {\itshape superposition}
of $n$ contributions $x^{\prime }( y_{i}) $ (scalar argument) for
$i=1,2,\cdots ,n$. Taken together, these constraints encourage the
formation of coding schemes in which independent subspaces are separately
coded, as required.

The constraints in Equation \ref{XRef-Equation-821165127} prevent
the full space of possible values of $\Pr ( y|x) $ or $x^{\prime
}( y) $ from being explored as $D$ is minimised, so they lead to
an {\itshape upper bound} $D_{1}+D_{2}$ on the FMC objective function
$D$ (i.e. $D\leq D_{1}+D_{2}$), which may be derived as \cite{Luttrell1997}
\begin{equation}
\begin{array}{rl}
 D_{1} & \equiv \frac{2}{n}\int dx \Pr ( x) \sum \limits_{y=1}^{M}\Pr
( y|x) \left| \left| x-x^{\prime }( y) \right| \right| ^{2} \\
 D_{2} & \equiv \frac{2\left( n-1\right) }{n}\int dx \Pr ( x) \left|
\left| x-\sum \limits_{y=1}^{M}\Pr ( y|x) x^{\prime }( y) \right|
\right| ^{2}
\end{array}%
\label{XRef-Equation-821175529}
\end{equation}

\noindent Note that $M$ (size of codebook) and $n$ (number of samples
drawn from codebook using $\Pr ( y|x) $) are effectively model order
parameters, whose values need to be chosen appropriately for each
encoder optimisation problem. The properties of the optimum solution
depend critically on the interplay between the statistical properties
of the training data and the model order parameters $M$ and $n$,
as will be seen in the simulations in Section \ref{XRef-Section-8211742}.
\subsection{Chains of Linked FMCs}\label{XRef-Subsection-821163750}
\begin{figure}[h]
\begin{center}
\includegraphics{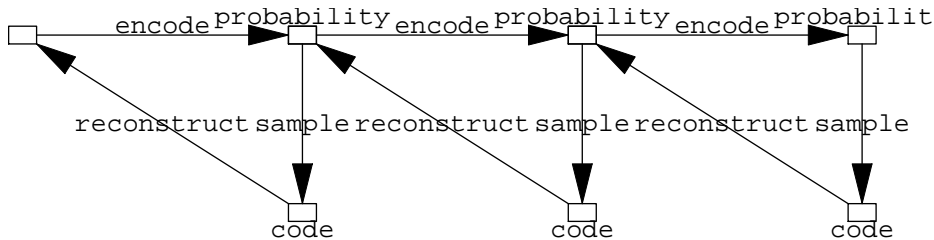}

\end{center}
\caption{A chain of linked FMCs, in which the output from each stage
is its vector of posterior probabilities (for all values of the
code index), which is then used as the input to the next stage.
Only 3 stages are shown, but any number may be used. More generally,
any acyclically linked network of FMCs may be used.}\label{XRef-FigureCaption-821165611}
\end{figure}

The FMC illustrated in Figure \ref{XRef-FigureCaption-82116438}
may be generalised to a chain of linked FMCs as shown in Figure
\ref{XRef-FigureCaption-821165611}. Each stage in this chain is
an FMC of the type shown in Figure \ref{XRef-FigureCaption-82116438},
and the vector of probabilities (for all values of the code index)
computed by each stage is used as the input vector to the next stage;
there are other ways of linking the stages together, but this is
the simplest possibility. The overall objective function is a weighted
sum of the FMC objective functions derived from each stage. The
total number of free parameters in an $L$ stage chain is $3L-1$,
which is the sum of 2 free parameters for each of the $L$ stages,
plus $L-1$ weighting coefficients; there are $L-1$ rather than $L$
weighting coefficients because the overall normalisation of the
objective function does not affect the optimum solution.

The chain of linked FMCs may be expressed mathematically by first
of all introducing an index $l$ to allow different stages of the
chain to be distinguished thus
\begin{equation}
\begin{array}{rl}
 M & \longrightarrow M^{\left( l\right) } \\
 x & \longrightarrow x^{\left( l\right) } \\
 y & \longrightarrow y^{\left( l\right) } \\
 x^{\prime } & \longrightarrow x^{\prime ( l) } \\
 n & \longrightarrow n^{\left( l\right) } \\
 D & \longrightarrow D^{\left( l\right) } \\
 D_{1} & \longrightarrow D_{1}^{\left( l\right) } \\
 D_{2} & \longrightarrow D_{2}^{\left( l\right) }
\end{array}
\end{equation}

\noindent The stages are then defined and linked together thus
\begin{equation}
\begin{array}{rl}
 x^{\left( l\right) } & \longrightarrow y^{\left( l\right) }\longrightarrow
x^{\prime ( l) } \\
 x^{\left( l+1\right) } & =\left( x_{1}^{\left( l+1\right) },x_{2}^{\left(
l+1\right) },\cdots ,x_{M^{\left( l\right) }}^{\left( l+1\right)
}\right)  \\
 x_{i}^{\left( l+1\right) } & =\Pr ( y^{\left( l\right) }=i|x^{\left(
l\right) }) , 1\leq i\leq M^{\left( l\right) }
\end{array}%
\label{XRef-Equation-8211800}
\end{equation}

\noindent The objective function and its upper bound are then given
by
\begin{equation}
\begin{array}{rl}
 D & =\sum \limits_{l=1}^{L}s^{\left( l\right) }D^{\left( l\right)
} \\
  & \leq D_{1}+D_{2} \\
  & =\sum \limits_{l=1}^{L}s^{\left( l\right) } \left( D_{1}^{\left(
l\right) }+D_{2}^{\left( l\right) }\right) 
\end{array}%
\label{XRef-Equation-821174456}
\end{equation}

\noindent where $s^{(l)}\geq 0$ is the weighting that is applied
to the contribution of stage $l$ of the chain to the overall objective
function.
\section{Simulations}\label{XRef-Section-8211742}

In this section the results of various simulations are presented,
which demonstrate some of the types of self-organising behaviour
exhibited by an encoder that consists of a chain of linked FMCs.
Synthetic, rather than real, training data are used in all of the
simulations, because this allows the basic types of behaviour to
be cleanly demonstrated.

In Section \ref{XRef-Subsection-821172357} the training data is
described. In Section \ref{XRef-Subsection-82117248} a single stage
encoder is trained on data that is a superposition of two randomly
positioned objects. In Section \ref{XRef-Subsection-821172424} this
is generalised to objects with correlated positions, and three different
types of behaviour are demonstrated: factorial encoding using both
a 1-stage and a 2-stage encoders (Section \ref{XRef-Subsection-821172441}),
joint encoding using a 1-stage encoder (Section \ref{XRef-Subsection-821172452}),
and invariant encoding using a 2-stage encoder (Section \ref{XRef-Subsection-82117252}).

In Appendix \ref{XRef-AppendixSection-821172518} the derivatives
of the objective function are derived, and in Appendix \ref{XRef-AppendixSection-82117268}
a gradient descent training algorithm based on these derivatives
is presented.
\subsection{Training Data}\label{XRef-Subsection-821172357}

The key property that this type of self-organising encoder exhibits
is its ability to automatically split up high-dimensional input
spaces into lower-dimensional subspaces, each of which is separately
encoded. This self-organisation manifests itself in many different
ways, depending on the interplay between the statistical properties
of the training data, and the 3 free parameters (i.e. the code book
size $M$, the number of code indices sampled $n$, and the stage
weighting $s$) per stage of the encoder (see Section \ref{XRef-Subsection-821163750}).

\begin{figure}[h]
\begin{center}
\includegraphics{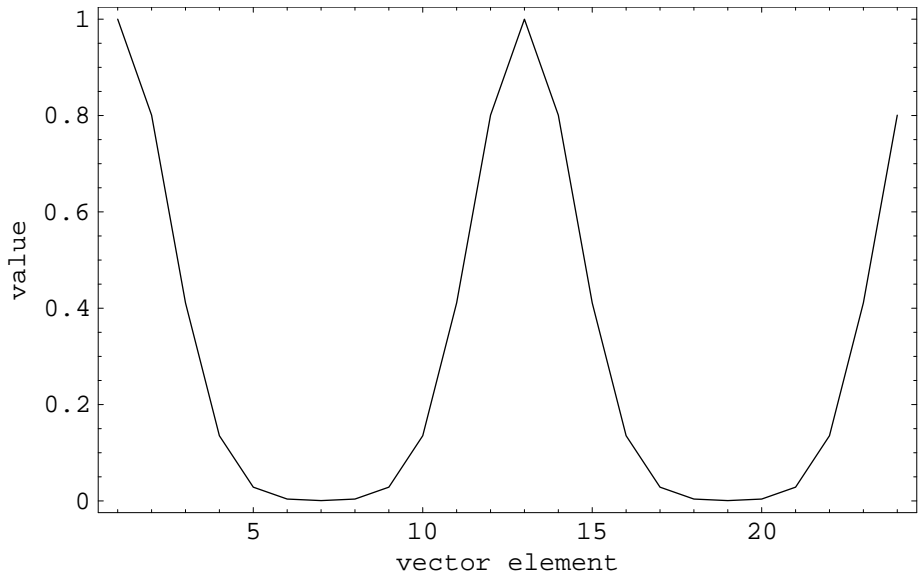}

\end{center}
\caption{An example of a typical training vector for $M=24$. Each
object is a Gaussian hump with a half-width of 1.5 units, and peak
amplitude of 1. The overall input vector is formed as a linear superposition
of the 2 objects. Note that the input vector is wrapped around circularly
to remove minor edge effects that would otherwise arise.}\label{XRef-FigureCaption-821172828}
\end{figure}

In order to demonstrate the various different basic types of self-organisation
it is necessary to use synthetic training data with controlled properties.
All of the types of self-organisation that will be demonstrated
in this paper may be obtained by training a 1-stage or 2-stage encoder
on 24-dimensional data (i.e. $M=24$) that consists of a superposition
of a pair of identical objects (with circular wraparound to remove
edge effects), such as is shown in Figure \ref{XRef-FigureCaption-821172828}.

In the simulations presented below, two different methods of selecting
the object positions are used: either the positions are statistically
independent, or they are correlated. In the independent case, each
object position is a random integer in the interval $[1,24]$. In
the correlated case, the first object position is a random integer
in the interval $[1,24]$, and the second object position is chosen
{\itshape relative to} the first one as an integer in the range
$[4,8]$, so that the mean object separation is 6 units.
\subsection{Independent Objects}\label{XRef-Subsection-82117248}

The simplest demonstration is to let a single stage encoder discover
the fact that the training data consists of a superposition of a
pair of objects, which is a type of independent component analysis
(ICA) \cite{Hyvarinen1999}. This may readily be done by setting
the parameter values as follows: code book size $M=16$, number of
code indices sampled $n=20$, $\varepsilon =0.2$ for 500 training
steps, $\varepsilon =0.1$ for a further 500 training steps.
\begin{figure}[h]
\begin{center}
\includegraphics{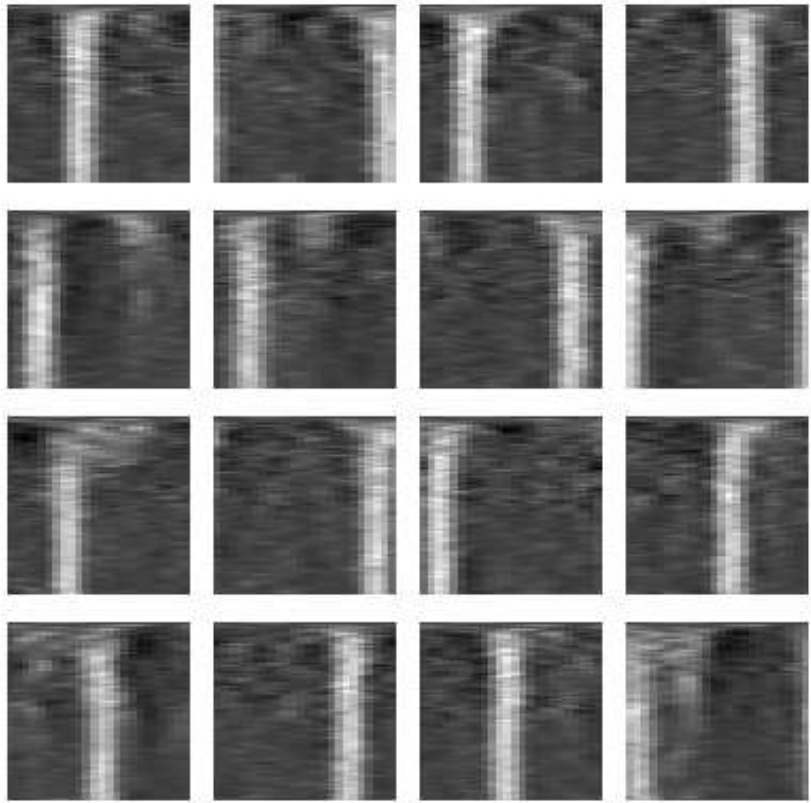}

\end{center}
\caption{A factorial encoder emerges when a single stage encoder
is trained on data that is a superposition of 2 objects in independent
locations.}\label{XRef-FigureCaption-82117315}
\end{figure}

The self-organisation of each of the 16 reconstruction vectors as
training progresses (measured down the page) is shown in Figure
\ref{XRef-FigureCaption-82117315}. After some initial confusion,
the reconstruction vectors self-organise so that each code index
corresponds to a {\itshape single} object at a well defined location,
whose width automatically adjusts itself so that the $M$ reconstruction
vectors cover the whole input space. This behaviour is non-trivial,
because each training vector is a superposition of a {\itshape pair}
of objects at independent locations, so two different code index
values must be sampled by the encoder (assuming that the two objects
are not at the same location); the relatively large choice $n=20$
ensures that it is highly likely that both code index values will
be amongst the $n$ random samples \cite{Luttrell1999a}. This result
is called a factorial encoder, because the objects are encoded separately.

The case of a joint encoder, where each code index corresponds to
a {\itshape pair} of objects at well defined locations, requires
a rather large code book when the objects are independent. However,
when correlations between the objects are introduced then the code
book can be reduced to a manageable size, as will be demonstrated
in the next section.
\subsection{Correlated Objects}\label{XRef-Subsection-821172424}

If the positions of the pair of objects are mutually correlated,
then they can be encoded in 3 fundamentally different ways:
\begin{enumerate}
\item Factorial encoder. This encoder ignores the correlations between
the objects, and encodes them as if they were 2 independent objects.
Each code index thus encodes a single object position, so many code
indices must be sampled in order to virtually guarantee that both
object positions are encoded \cite{Luttrell1999a}. This result is
a type of independent component analysis (ICA) \cite{Hyvarinen1999}.
\item Joint encoder. This encoder regards each possible joint placement
of the 2 objects as a distinct configuration. Each code index thus
encodes a pair of object positions, so only one code index needs
to be sampled in order to guarantee that both object positions are
encoded \cite{Luttrell1999a}. This result is basically the same
as what would be obtained by using a standard VQ \cite{LindeBuzoGray1980}.
\item Invariant encoder. This encoder regards each possible placement
of the centroid of the 2 objects as a distinct configuration, but
regards all possible object separations (for a given centroid) as
being equivalent. Each code index thus encodes only the centroid
of the pair of objects. This type of encoder does not arise when
the objects are independent. This is similar to self-organising
transformation invariant detectors described in \cite{Webber1994}.
\begin{figure}[h]
\begin{center}
\includegraphics{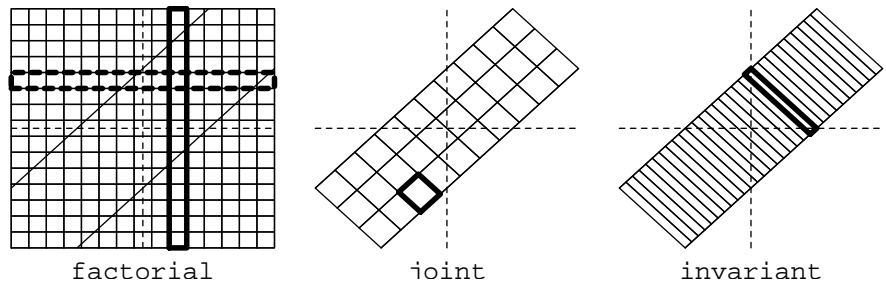}

\end{center}
\caption{Three alternative ways of using 30 code indices to encode
a pair of correlated variables. The typical code cells are shown
in bold.}\label{XRef-FigureCaption-821173851}
\end{figure}
\end{enumerate}

Each of these 3 possibilities is shown in Figure \ref{XRef-FigureCaption-821173851},
where the diagrams are meant only to be illustrative. The correlated
variables live in the large 2-dimensional rectangular region extending
from bottom-left to top-right of each diagram.

The factorial encoder has two orthogonal sets of long thin rectangular
code cells, and the diagram shows how a pair of such cells intersect
to define a small square code cell. The joint encoder behaves as
a standard vector quantiser, and is illustrated as having a set
of square code cells, although their shapes will not be as simple
as this in practice. The invariant encoder ideally has a set of
long thin rectangular code cells that encode only the long diagonal
dimension.

In all 3 cases there is overlap between code cells. In the case
of the factorial and joint encoders the overlap tends to be only
between nearby code cells, whereas in the case of an invariant encoder
the range of the overlap is usually much greater, as will be seen
in the numerical simulations below. In practice the optimum encoder
may not be a clean example of one of the types illustrated in Figure
\ref{XRef-FigureCaption-821173851}, as will also be seen in the
numerical simulations below.
\subsection{Factorial Encoding}\label{XRef-Subsection-821172441}

A factorial encoder may be trained by setting the parameter values
as follows: code book size $M=16$, number of code indices sampled
$n=20$, $\varepsilon =0.2$ for 500 training steps, $\varepsilon
=0.1$ for a further 500 training steps.
\begin{figure}[h]
\begin{center}
\includegraphics{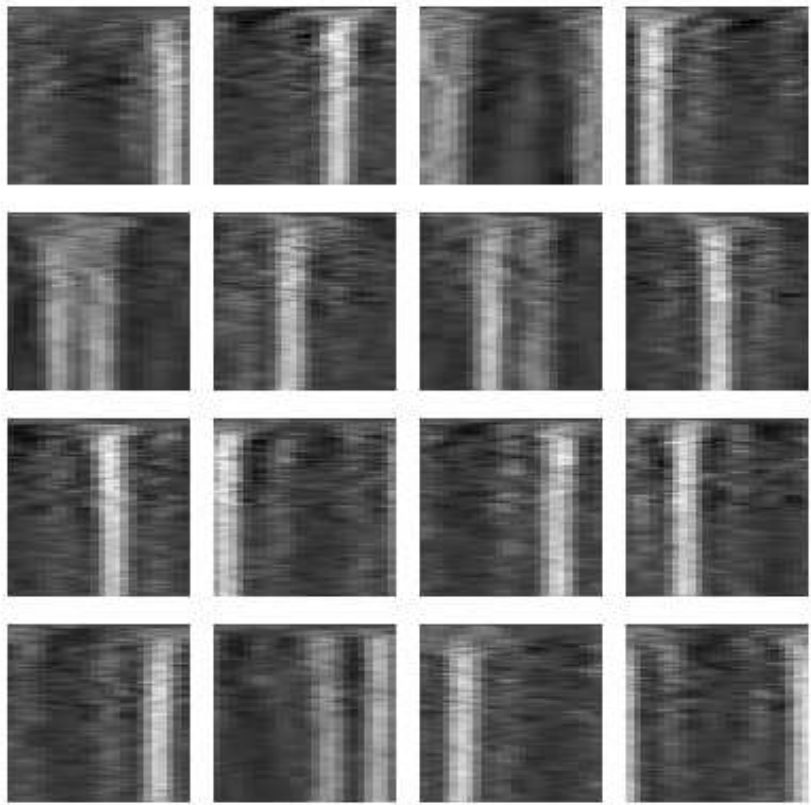}

\end{center}
\caption{A factorial encoder emerges when a single stage encoder
is trained on data that is a superposition of 2 objects in correlated
locations.}\label{XRef-FigureCaption-821174439}
\end{figure}

The result is shown in Figure \ref{XRef-FigureCaption-821174439}
which should be compared with the result for independent objects
in Figure \ref{XRef-FigureCaption-82117315}. The presence of correlations
degrades the quality of this factorial code relative to the case
of independent objects. The contamination of the factorial code
takes the form of a few code indices which respond jointly to the
pair of objects.

The joint coding contamination of the factorial code can be reduced
by using a 2-stage encoder, in which the second stage has the same
values of $M$ and $n$ as the first stage (although identical parameter
values are not necessary), and (in this case) both stages have the
same weighting in the objective function (see Equation \ref{XRef-Equation-821174456}).
\begin{figure}[h]
\begin{center}
\includegraphics{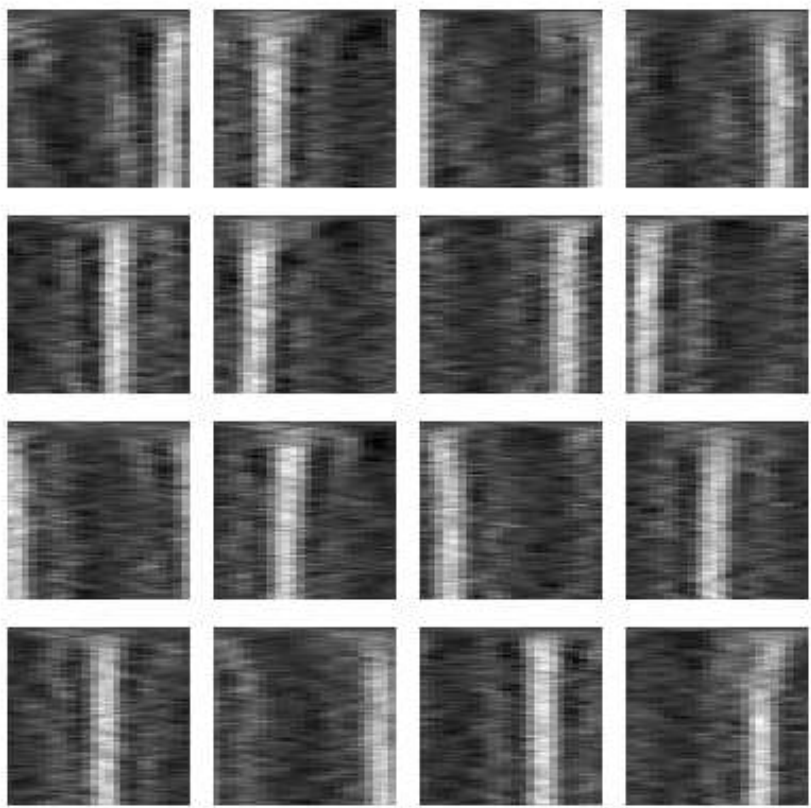}

\end{center}
\caption{The factorial encoder is improved, by the removal of the
joint encoding contamination, when a 2-stage encoder is used.}\label{XRef-Figure-821174614}
\end{figure}

The results are shown in Figure \ref{XRef-Figure-821174614}. The
reason that the second stage encourages the first to adopt a pure
factorial code is quite subtle. The result shown in Figure \ref{XRef-Figure-821174614}
will lead to the first stage producing an output in which 2 code
indices (one for each object) each typically have probability $\frac{1}{2}$
of being sampled, and all of the remaining code indices have a very
small probability (this is an approximation which ignores the fact
that the code cells overlap). On the other hand, Figure \ref{XRef-FigureCaption-821174439}
will lead to an output in which the probability can be concentrated
on a single code index, if it can jointly code the pair of objects.
However, the contribution of the second stage to the overall objective
function encourages it to encode the vector of probabilities output
by the first stage with minimum Euclidean reconstruction error,
which is easier to do if the situation is as in Figure \ref{XRef-Figure-821174614}
rather than as in Figure \ref{XRef-FigureCaption-821174439}. In
effect, the second stage likes to see an output from the first stage
in which more than one code index has a significant probability
of being sampled, which favours factorial coding over joint encoding.
\subsection{Joint Encoding}\label{XRef-Subsection-821172452}

A joint encoder may be trained by setting the parameter values as
follows: code book size $M=16$, number of code indices sampled $n=3$,
$\varepsilon =0.2$ for 500 training steps, $\varepsilon =0.1$ for
a further 500 training steps, $\varepsilon =0.05$ for a further
1000 training steps. This is the same as the parameter values for
the factorial encoder above, except that $n$ has been reduced to
$n=3$, and the training schedule has been extended.
\begin{figure}[h]
\begin{center}
\includegraphics{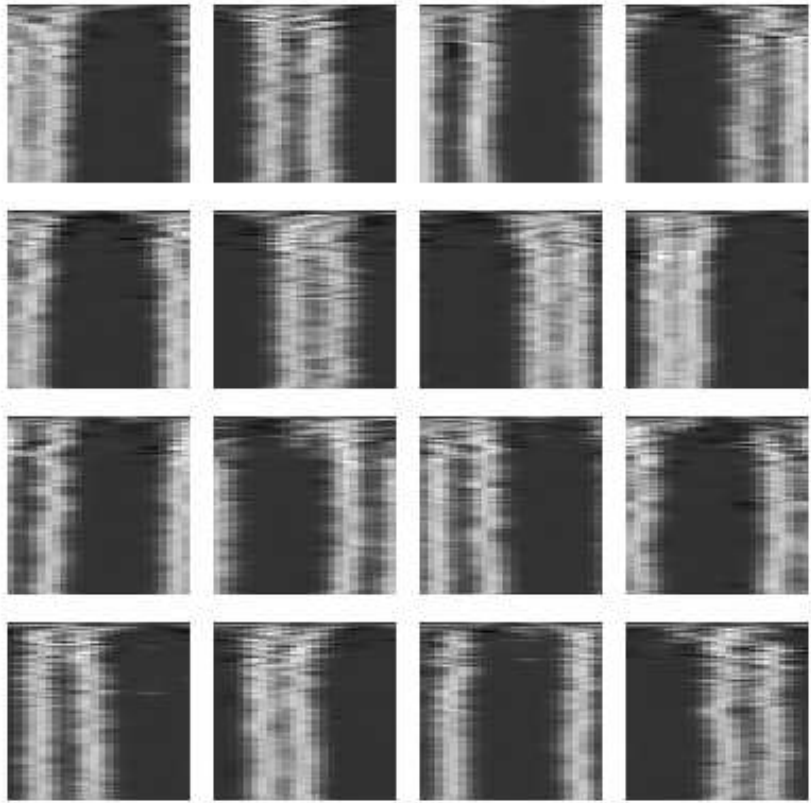}

\end{center}
\caption{A joint encoder emerges when a single stage encoder is
trained on data that is a superposition of 2 objects in correlated
locations.}\label{XRef-Figure-821174852}
\end{figure}

The result is shown in Figure \ref{XRef-Figure-821174852}. After
some initial confusion, the reconstruction vectors self-organise
so that each code index corresponds to a {\itshape pair} of objects
at well defined locations, so the code index jointly encodes the
pair of object positions; this is a joint encoder. The small value
of $n$ prevents a factorial encoder from emerging \cite{Luttrell1999a}.
\subsection{Invariant Encoding}\label{XRef-Subsection-82117252}

An invariant encoder may be trained by using a 2-stage encoder,
and setting the parameter values identically in each stage as follows
(where the weighting of the second stage relative to the first is
denoted as $s$): code book size $M=16$, number of code indices sampled
$n=3$, $\varepsilon =0.2$ and $s=5$ for 500 training steps, $\varepsilon
=0.1$ and $s=10$ for a further 500 training steps, $\varepsilon
=0.05$ and $s=20$ for a further 500 training steps, $\varepsilon
=0.05$ and $s=40$ for a further 500 training steps. This is basically
the same as the parameter values used for the joint encoder above,
except that there are now 2 stages, and the weighting of the second
stage is progressively increased throughout the training schedule.
Note that the large value that is used for $s$ is offset to a certain
extent by the fact that the ratio of the normalisation of the inputs
to the first and second stages is very large; the anomalous normalisation
of the input to the first stage could be removed by insisting that
the input to the first stage is a vector of probabilities, but that
is not done in these simulations.
\begin{figure}[h]
\begin{center}
\includegraphics{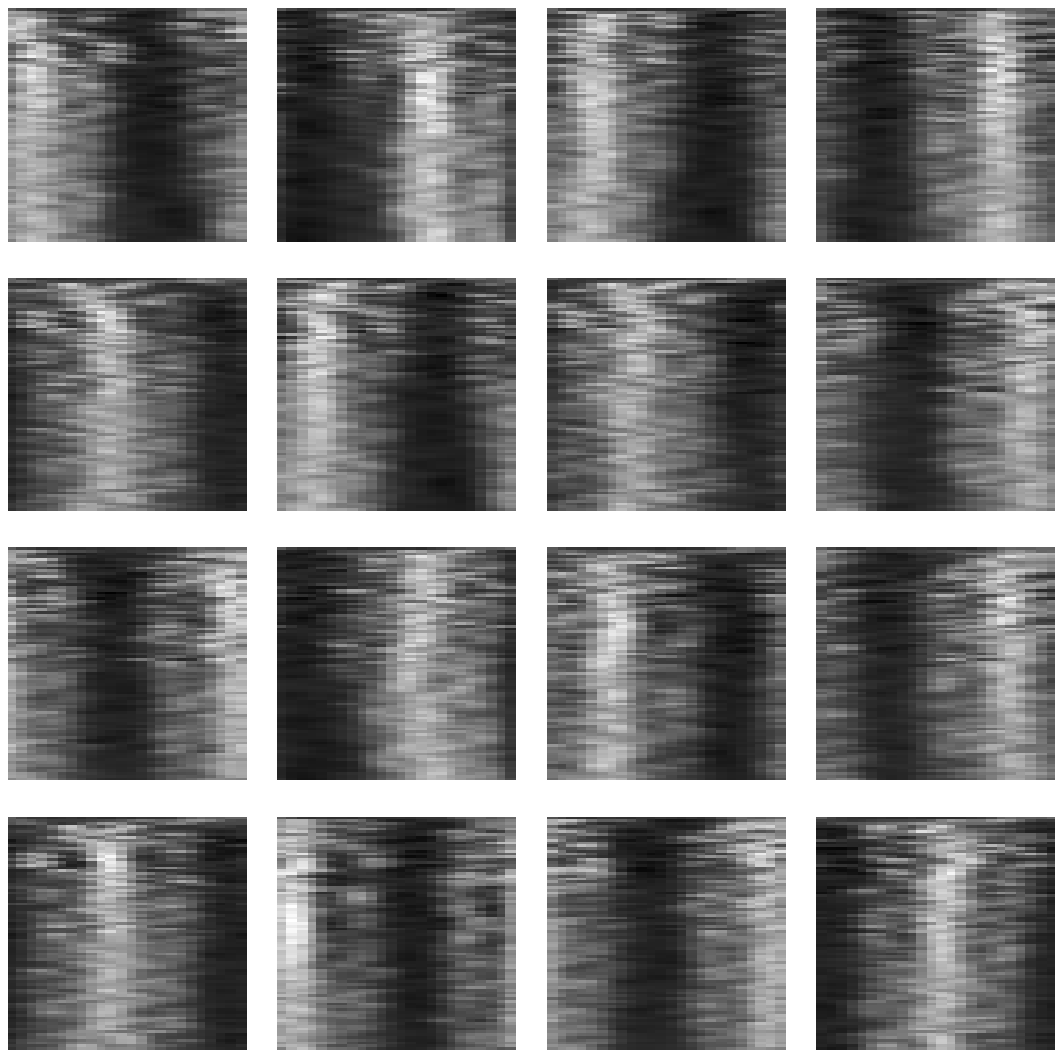}

\end{center}
\caption{An invariant encoder emerges when 2-stage encoder is trained
on data that is a superposition of 2 objects in correlated locations.}\label{XRef-Figure-82117519}
\end{figure}

The result is shown in Figure \ref{XRef-Figure-82117519}. During
the early part of the training schedule the weighting of the second
stage is still relatively small, so it has the effect of turning
what would otherwise have been a joint encoder into a factorial
encoder; this is analogous to the effect observed when Figure \ref{XRef-FigureCaption-821174439}
becomes Figure \ref{XRef-Figure-821174614}. However, as the training
schedule progresses the weighting of the second stage increases
further, and the reconstruction vectors self-organise so that each
code index corresponds to a {\itshape pair} of objects with a well
defined centroid but indeterminate separation. Thus each code index
encodes only the centroid of the pair of objects and ignores their
separation. This is a new type of encoder that arises when the objects
are correlated, and it will be called an {\itshape invariant} encoder,
in recognition of the fact that its output is invariant with respect
to the separation of the objects.

Note that in these results there is a large amount of overlap between
the code cells, which should be taken into account when interpreting
the illustration in Figure \ref{XRef-FigureCaption-821173851}. This
is an extreme example the second stage preferring an output from
the first stage in which more than one code index has a significant
probability of being sampled; the large amount of overlap between
code cells means that many code indices have a significant probability
of being sampled.
\section{Conclusions}

The numerical results presented in this paper show that a stochastic
vector quantiser (SVQ) can self-organise to find a variety of different
types of way of encoding high-dimensional input vectors. Three fundamentally
different types of encoder have been demonstrated, which differ
in the way that they build a reconstruction that approximates the
input vector:

1. A factorial encoder uses a reconstruction that is superposition
of a {\itshape number} of vectors that each lives in a well defined
input subspace, which is useful for discovering constituent objects
in the input vector. This result is a type of independent component
analysis (ICA) \cite{Hyvarinen1999}.

2. A joint encoder uses a reconstruction that is a {\itshape single}
vector that lives in the whole input space. This result is basically
the same as what would be obtained by using a standard VQ \cite{LindeBuzoGray1980}.

3. An invariant encoder uses a reconstruction that is a {\itshape
single} vector that lives in a subspace of the whole input space,
so it ignores some dimensions of the input vector, which is therefore
useful for discovering correlated objects whilst rejecting uninteresting
fluctuations in their relative coordinates. This is similar to self-organising
transformation invariant detectors described in \cite{Webber1994}.

More generally, the encoder will be a hybrid of these basic types,
depending on the interplay between the statistical properties of
the input vector and the parameter settings of the SVQ.

\appendix

\section{Derivatives of the Objective Function}\label{XRef-AppendixSection-821172518}

In order to minimise $D_{1}+D_{2}$ it is necessary to compute its
derivatives. The derivatives were presented in detail in \cite{Luttrell1997}
for a single stage chain (i.e. a single FMC). The purpose of this
appendix is to extend this derivation to a multi-stage chain of
linked FMCs. In order to write the various expressions compactly,
infinitesimal variations will be used thoughout this appendix, so
that $\delta ( u v) =\delta u v+u \delta v$ will be written rather
than $\frac{\partial (u v)}{\partial \theta }=\frac{\partial u}{\partial
\theta }v+u\frac{\partial v}{\partial \theta }$ (for some parameter
$\theta $). The calculation will be done in a top-down fashion,
differentiating the objective function first, then differentiating
anything that the objective function depends on, and so on following
the dependencies down until only constants are left (this is essentially
the chain rule of differentiation).

The derivative of $D_{1}+D_{2}$ (defined in Equation \ref{XRef-Equation-821174456})
is given by
\begin{equation}
\delta \sum \limits_{l=1}^{L}s^{\left( l\right) } \left( D_{1}^{\left(
l\right) }+D_{2}^{\left( l\right) }\right) =\sum \limits_{l=1}^{L}s^{\left(
l\right) } \left( \delta D_{1}^{\left( l\right) }+\delta D_{2}^{\left(
l\right) }\right) 
\end{equation}

\noindent The derivatives of the $D_{1}^{(l)}$ and $D_{2}^{(l)}$
parts (defined in Equation \ref{XRef-Equation-821175529}, with appropriate
$(l)$ superscripts added) of the contribution of stage $l$ to $D_{1}+D_{2}$
are given by (dropping the $(l)$ superscripts again, for $\operatorname{clarity})$
\begin{equation}
\begin{array}{rl}
 \delta D_{1} & =\frac{2}{n}\int dx \Pr ( x) \sum \limits_{y=1}^{M}\left(
\delta Pr( y|x) \left| \left| x-x^{\prime }( y) \right| \right|
^{2}\right.  \\
   & \left. +2\Pr ( y|x) \left( \delta x-\delta x^{\prime }( y)
\right) \right) .\left( x-x^{\prime }( y) \right)  \\
 \delta D_{2} & =\frac{4\left( n-1\right) }{n}\int dx \Pr ( x) \left(
\delta x-\sum \limits_{y=1}^{M}\left( \delta Pr( y|x) x^{\prime
}( y) \right. \right.  \\
   & \left. \left. +\Pr ( y|x) \delta x^{\prime }( y) \right) \right)
.\left( x-\sum \limits_{y^{\prime }=1}^{M}\Pr ( y^{\prime }|x) x^{\prime
}( y^{\prime }) \right) 
\end{array}%
\label{XRef-Equation-821175930}
\end{equation}

The first step in modelling $\Pr ( y|x) $ is to explicitly state
the fact that it is a probability, which is a non-negative normalised
quantity. This may be done as follows
\begin{equation}
\Pr ( y|x) =\frac{Q( y|x) }{\sum \limits_{y^{\prime }=1}^{M}Q( y^{\prime
}|x) }
\end{equation}

\noindent where $Q( y|x) \geq 0$. The $Q( y|x) $ are unnormalised
probabilities, and $\sum \limits_{y^{\prime }=1}^{M}Q( y^{\prime
}|x) $ is the normalisation factor. The derivative of $\Pr ( y|x)
$ is given by
\begin{equation}
\frac{\delta Pr( y|x) }{\Pr ( y|x) }=\frac{1}{Q( y|x) }\left( \delta
Q( y|x) -\Pr ( y|x) \sum \limits_{y^{\prime }=1}^{M}\delta Q( y^{\prime
}|x) \right) 
\end{equation}

The second step in modelling $\Pr ( y|x) $ is to introduce an explicit
parameteric form for $Q( y|x) $. The following sigmoidal function
will be used in this paper
\begin{equation}
Q( y|x) =\frac{1}{1+\exp ( -w( y) .x-b( y) ) }
\end{equation}

\noindent where $w( y) $ is a weight vector and $b( y) $ is a bias.
The derivative of $Q( y|x) $ is given by
\begin{equation}
\delta Q( y|x) =Q( y|x) \left( 1-Q( y|x) \right) \left( \delta w(
y) .x+w( y) .\delta x+\delta b( y) \right) %
\label{XRef-Equation-821175938}
\end{equation}

This has reduced the $\delta D_{1}$ and $\delta D_{2}$ derivatives
to $\delta w( y) $, $\delta b( y) $, $\delta x^{\prime }( y) $ and
$\delta x$ derivatives. The $\delta w( y) $, $\delta b( y) $ and
$\delta x^{\prime }( y) $ derivatives relate directly to the parameters
being optimised and thus need no further simplification, however
the $\delta x$ derivatives in Equation \ref{XRef-Equation-821175930}
and Equation \ref{XRef-Equation-821175938} need some further attention.
The $\delta x$ derivative arises only in multi-stage chains of FMCs,
and because of the way in which stages of the chain are linked together
(see Equation \ref{XRef-Equation-8211800}) it is equal to the derivative
of the vector of probabilities output by the previous stage. Thus
the $\delta x$ derivative may be obtained by following its dependencies
back through the stages of the chain until the first layer is reached;
this is essentially the chain rule of differentiation. This ensures
that for each stage the partial derivatives include the additional
contributions that arise from forward propagation through later
stages, as described in Appendix \ref{XRef-AppendixSection-82117268}.

\section{Training Algorithm}\label{XRef-AppendixSection-82117268}

Assuming that $\Pr ( y|x) $ is modelled as in appendix A (i.e. $\Pr
( y|x) =\frac{Q( y|x) }{\sum \limits_{y^{\prime }=1}^{M}Q( y^{\prime
}|x) }$ and $Q( y|x) =\frac{1}{1+\exp ( -w( y) .x-b( y) ) }$), then
the partial derivatives of $D_{1}+D_{2}$ with respect to the 3 types
of parameters in a single stage of the encoder may be denoted as
\begin{equation}
\begin{array}{rl}
 g_{w}( y)  & \equiv \frac{\partial \left( D_{1}+D_{2}\right) }{\partial
w( y) } \\
 g_{b}( y)  & \equiv \frac{\partial \left( D_{1}+D_{2}\right) }{\partial
b( y) } \\
 g_{x}( y)  & \equiv \frac{\partial \left( D_{1}+D_{2}\right) }{\partial
x^{\prime }( y) }
\end{array}
\end{equation}

\noindent This may be generalised to each stage of a multi-stage
encoder by including an $(l)$ superscript, and ensuring that for
each stage the partial derivatives include the additional contributions
that arise from forward propagation through later stages; this is
essentially an application of the chain rule of differentiation,
using the derivatives $\frac{\partial x^{(l+1)}}{\partial w^{(l)}(
y^{(l)}) }$ and $\frac{\partial x^{(l+1)}}{\partial b^{(l)}( y^{(l)})
}$ to link the stages together (see appendix A).

A simple algorithm for updating these parameters is (omitting the
$(l)$ superscript, for clarity)
\begin{equation}
\begin{array}{rl}
 w( y)  & \longrightarrow w( y) -\varepsilon \frac{g_{w}( y) }{g_{w,0}}
\\
 b( y)  & \longrightarrow b( y) -\varepsilon \frac{g_{b}( y) }{g_{b,0}}
\\
 x^{\prime }( y)  & \longrightarrow x^{\prime }( y) -\varepsilon
\frac{g_{x}( y) }{g_{x,0}}
\end{array}
\end{equation}

\noindent where $\varepsilon $ is a small update step size parameter,
and the three normalisation factors are defined as
\begin{equation}
\begin{array}{rl}
 g_{w,0} & \equiv \begin{array}{c}
 \max  \\
 y
\end{array}\sqrt{\frac{\left| \left| g_{w}( y) \right| \right| ^{2}}{\dim
x}} \\
 g_{b,0} & \equiv \begin{array}{c}
 \max  \\
 y
\end{array}\left| b( y) \right|  \\
 g_{x,0} & \equiv \begin{array}{c}
 \max  \\
 y
\end{array}\sqrt{\frac{\left| \left| g_{x}( y) \right| \right| ^{2}}{\dim
x}}
\end{array}
\end{equation}

The $\frac{g_{w}( y) }{g_{w,0}}$ and $\frac{g_{x}( y) }{g_{x,0}}$
factors ensure that the maximum update step size for $w( y) $ and
$x^{\prime }( y) $ is $\varepsilon  \dim  x$ (i.e. $\varepsilon
$ per dimension), and the $\frac{g_{b}( y) }{g_{b,0}}$ factor ensures
that the maximum update step size for $b( y) $ is $\varepsilon $.
When a stationary point of $D_{1}+D_{2}$ is reached, the finite
size of $\varepsilon $ prevents the parmater values from converging
to a perfectly stationary solution, and instead they jump around
in its neighbourhood.

This update algorithm can be generalised to use a different $\varepsilon
$ for each stage of the encoder, and also to allow a different $\varepsilon
$ to be used for each of the 3 types of parameter. Furthermore,
the size of $\varepsilon $ can be varied as training proceeds, usually
starting with a large value, and then gradually reducing its size
to obtain an accurate estimate of the stationary solution. It is
not possible to give general rules for exactly how to do this, because
training conditions depend very much on the statistical properties
of the training set.\label{TitleNotes}

\end{document}